\pdfoutput=1

\documentclass[11pt]{article}

\usepackage[final]{acl}

\usepackage{times}
\usepackage{latexsym}
\usepackage{amsmath}
\usepackage{amssymb}
\usepackage{dsfont}
\usepackage[ruled,vlined]{algorithm2e}
\usepackage[T1]{fontenc}

\usepackage[utf8]{inputenc}

\usepackage{microtype}

\usepackage{inconsolata}
\usepackage{colortbl}
\usepackage{float}

\usepackage{graphicx}

\usepackage{contour}
\usepackage{booktabs}

\newcommand{\hide}[1]{}

\newcommand{\AHAMZAUP}{{\^{A}}}

\newcommand{\TAMAR}{{$\hbar$}}

\newcommand{\SHIN}{{\v{s}}}
\newcommand{\AYN}{{$\varsigma$}}
\newcommand{\GAYN}{{$\gamma$}}

\newcommand{\SHADDA}{{$\sim$}}

\usepackage{arabtex}

\usepackage{utf8}
\usepackage{epstopdf}

\newcommand{\phon}{ph}
\newcommand{\ortho}{or}
\newcommand{\etym}{et}

\title{The Arabic Generality Score:\\Another Dimension of Modeling Arabic Dialectness}

\author{Sanad Shaban,\textsuperscript{1} Nizar Habash\textsuperscript{1,2} \\  \textsuperscript{1}MBZUAI, \textsuperscript{2}New York University Abu Dhabi\\
  \texttt{sanad.shaban@mbzuai.ac.ae, nizar.habash@nyu.edu}
  }

\begin{document}
\maketitle

\setcode{utf8}
\vocalize 


\begin{abstract}

Arabic dialects form a diverse continuum, yet NLP models often treat them as discrete categories. Recent work addresses this issue by modeling dialectness as a continuous variable, notably through the Arabic Level of Dialectness (ALDi). However, ALDi reduces complex variation to a single dimension.
We propose a complementary measure: the Arabic Generality Score (AGS), which quantifies how widely a word is used across dialects. We introduce a pipeline that combines word alignment, etymology-aware edit distance, and smoothing to annotate a parallel corpus with word-level AGS. A regression model is then trained to predict AGS in context.
Our approach outperforms strong baselines, including state-of-the-art dialect ID systems, on a multi-dialect benchmark. AGS offers a scalable, linguistically grounded way to model lexical generality, enriching representations of Arabic dialectness.

\end{list} 
\end{abstract}

\section{Introduction}

%
%
Arabic exhibits a well-known case of diglossia: Modern Standard Arabic (MSA) functions as the high variety in formal settings, while a range of Dialectal Arabic (DA) varieties are used in everyday communication \cite{ferguson1959diglossia}. These dialects differ significantly from MSA, and from each other, in vocabulary, phonology, morphology, and syntax, often resulting in limited mutual intelligibility. 
And rather than forming discrete categories, they exist on a continuum. Token-level dissimilarity between MSA and dialects ranges from 37\% to 67\% \cite{fine-grained-DI-madar}, and speakers frequently blend MSA and DA depending on context and background \cite{badawi1973mustawayat, badawi1986dictionary,diez-white-arabic}. Dialects themselves also share features and borrow from other languages, such as French and English \cite{hamed-etal-2020-arzen}.

\begin{figure}[t]
\centering
\includegraphics[width=0.95\columnwidth]{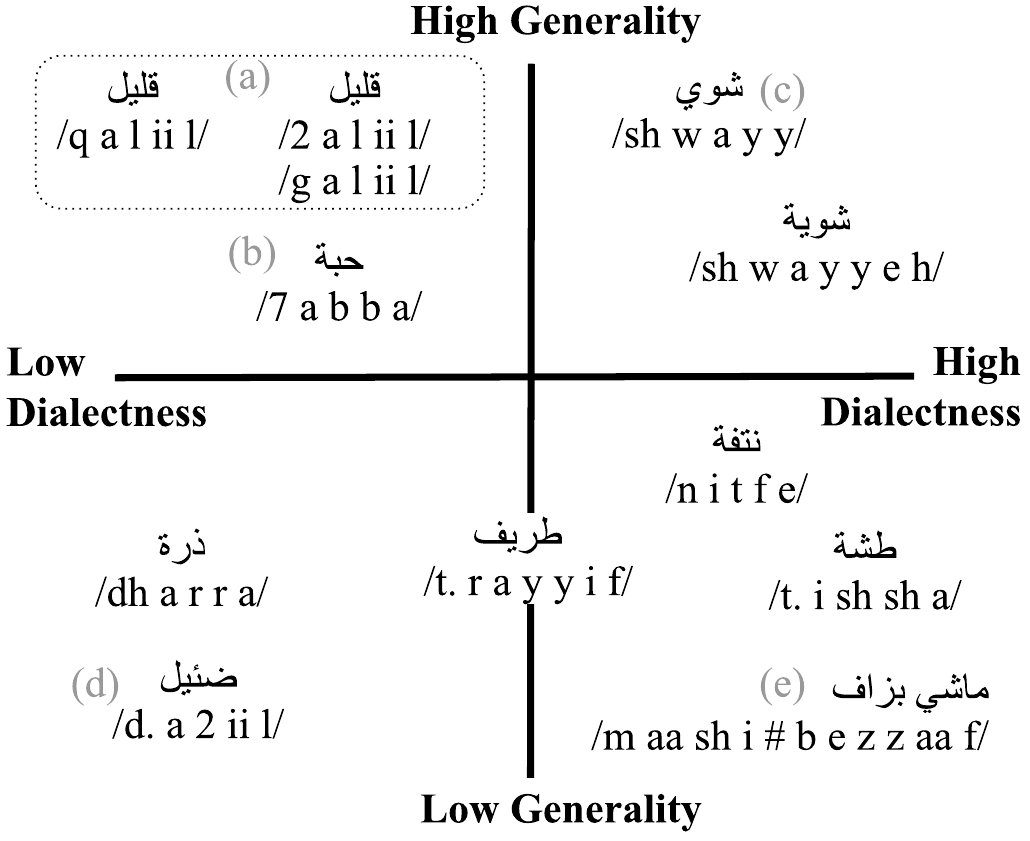}
\caption{Different MSA and DA words and their CAPHI phonology representation, with varying generality and dialectness levels; all mean `(a) little (bit).'}
\label{fig:ALDiXAGS}
\end{figure}

Most computational models still frame dialect identification (DID) as a fixed-label classification task \cite{DI-long-paper-zaidan-2014, fine-grained-DI-madar, NADI-2020, NADI-2021, NADI-2022, NADI-2023}, overlooking intra-dialect variation and cross-dialectal overlap. More recent approaches, like ALDi \cite{ALDI}, model dialectness as a continuous scalar, but reduce diverse signals to a single dimension.
%
%

We propose the Arabic Generality Score (AGS), a complementary dimension to ALDi that captures how broadly a word is used across dialects.
Figure~\ref{fig:ALDiXAGS} illustrates how AGS enables distinctions between widely shared and highly localized dialectal items, even when both diverge from MSA. Together, ALDi and AGS define a two-dimensional space for modeling Arabic dialectness.
We introduce a novel pipeline to annotate a parallel dialect corpus with word-level AGS, combining alignment, etymology- and phonology-aware edit distance, and smoothing. We then fine-tune a BERT-based regression model to predict AGS in context. Evaluation on multi-dialect data shows that our model outperforms strong baselines and captures lexical generality more effectively.

%


\section{Related Work}
\label{sec:relatedwork}

\subsection{Arabic Dialect Identification}

Arabic dialect identification (DID) has traditionally been treated as a single-label classification task, assigning each text a discrete dialect label such as Egyptian, Levantine, or Gulf Arabic, or Modern Standard Arabic (MSA) \cite{DI-long-paper-zaidan-2014}. Later work introduced finer-grained setups, including city-level classification with the MADAR-26 corpus \cite{madar-corpus-and-lexicon, fine-grained-DI-madar}. However, such approaches assume dialectal boundaries are clean, despite evidence of overlap and mutual influence among varieties.

Recent work highlights the limitations of discrete labeling. Texts often exhibit features from multiple dialects and MSA, especially in social media. Error analyses show that many supposed misclassifications are linguistically plausible alternatives \cite{limitations-of-single-label-DI, madar_error_analysis}, motivating multi-label and continuous approaches. The Arabic Level of Dialectness (ALDi) framework \cite{ALDI}, for instance, models dialectness on a continuous scale rather than as a hard class boundary.

This shift is also evident in shared tasks. The NADI series evolved from strict classification in 2020 to multi-label and continuous subtasks by 2024 \cite{NADI-2020, NADI-2021, NADI-2024}, including tasks for ALDi estimation. These developments reflect a growing consensus: effective DID must account for the fluid, gradient nature of Arabic dialects, paving the way for multidimensional frameworks such as ours.

\subsection{Dialectness} 

Dialectness refers to how much a text diverges from Modern Standard Arabic (MSA) and exhibits features of Dialectal Arabic (DA). Sociolinguistically, Badawi’s five-level model \cite{badawi1973mustawayat} captures this as a continuum from Classical Arabic to purely colloquial speech. It reflects how speakers shift registers based on context and education.

Computationally, early approaches measured dialectness using word frequency ratios in DA versus MSA corpora \cite{DI-long-paper-zaidan-2014, Arabench-DA-Eng-MT-benchmark}. Later work introduced more detailed annotation schemes labeling tokens and segments by their deviation from MSA, incorporating orthographic, morphological, and lexical cues \cite{dialectness_annotation_guidelines}. While accurate, these methods are labor-intensive and require expert annotators.

Crowdsourced alternatives, such as the Arabic Online Commentary (AOC) corpus \cite{AOC}, used coarser labels (e.g., “Little” or “Mostly Dialectal”) but lacked consistent guidelines. Building on these efforts, the Arabic Level of Dialectness (ALDi) framework \cite{ALDI} introduced continuous sentence-level scores based on averaged crowd annotations. A regression model (Sentence-ALDi) fine-tuned on MarBERT predicts these scores on a 0 to 1 scale. ALDi captures a fine-grained, replicable measure of dialectal divergence. However, it remains a single-axis metric and may conflate different types of variation, such as geographically localized versus widely used forms. 

To address this limitation, we introduce the Arabic Generality Score (AGS), a complementary axis that captures how widely a word is used across dialects, independent of its divergence from MSA. AGS allows us to distinguish between regionally specific and broadly shared dialectal forms, information that ALDi alone cannot provide. Together, AGS and ALDi define a two-dimensional space of variation, offering a richer and more interpretable model of the Arabic dialect continuum.

\section{Linguistic Background}
\label{sec:ling}

\subsection{Rich Morphology and Noisy Orthography}
\label{subsec:DA_challenges}

Arabic presents significant NLP challenges due to its rich morphology and highly inconsistent orthography, especially in dialects. Root-and-pattern morphology, combined with affixes and clitics, leads to high sparsity, while the lack of diacritics increases ambiguity. Dialects further complicate processing with informal spelling: a single word may appear in over 25+ forms \cite{Eskander:2013:processing,codafication, Habash:2018:unified}. In the absence of standardized conventions, spelling choices vary based on pronunciation and etymology. Writers may opt for phonetic spellings that reflect local speech or etymological ones that preserve MSA roots; sometimes these align, but often they diverge.
For instance, consider the word for `heart', \<قلب>~\textit{qlb}\footnote{HSB Arabic transliteration \cite{Habash:2007:arabic-transliteration}.}~/qalb/ in MSA. It is pronounced /2alb/ in Beirut, replacing the /q/ with a glottal stop /2/. A person from Beirut may choose to spell it phonetically as \<ألب>~\textit{Âlb}, or etymologically, as \<قلب>~\textit{qlb}.

\subsection{CODA and CAPHI}




We adopt a normalization framework designed to reduce surface variation and enhance cross-dialect comparability \cite{Habash:2018:unified}. 

The Conventional Orthography for Dialectal Arabic (\textbf{CODA}) defines a consistent spelling system for dialects using the Arabic script. It reduces sparsity by mapping surface forms to their etymological roots where appropriate. In the example from Section~\ref{subsec:DA_challenges}, CODA maps \<ألب>~\textit{Âlb} to \<قلب>~\textit{qlb}, recognizing them as orthographic variants of the same underlying word.

The CAMeL Arabic Phonetic Inventory (\textbf{CAPHI}) provides a complementary phonological transcription layer specifically designed for Arabic. It captures dialect-specific pronunciations of CODA-normalized words. For instance, \<قلب>~\textit{q-l-b} may be realized as /qalb/, /2alb/, or /galb/ in MSA, BEI, and DOH, respectively.

Together, CODA and CAPHI expose structural equivalence across dialects and support normalization-aware approaches to dialect modeling.

\section{The Arabic Generality Score}
\label{sec:agsframework}
We introduce a new dimension of Arabic dialectness: the \textbf{Arabic Generality Score (AGS)}~$\in [0,1]$, which quantifies the extent to which an Arabic utterance is used across Arabic dialects and MSA. A higher AGS indicates broader generality across dialects and MSA, while a lower AGS reflects specificity. Although AGS may correlate with ALDi in parts of the dialect-MSA space---particularly where dialects intersect with MSA---it captures a distinct phenomenon. Many features with high AGS are not necessarily close to MSA. For example, dialectal terms such as \<ليش>~\textit{lyš}~`why', \<مافي>~\textit{mAfy}~`there isn’t', and \<شوي>~\textit{šwy{\SHADDA}}~`a little' are widespread across Levantine, Gulf, and North African varieties. These items are perceived as highly \textit{general} due to their cross-dialectal prevalence, despite their non-standard status. Conversely, expressions like \<السلام عليكم>~\textit{AlslAm {\AYN}lykm}~`Hello' are standard but also frequent in dialectal contexts. As illustrated in Figure~\ref{fig:ALDiXAGS}, words span the full space defined by AGS and ALDi.

In the next two sections, we introduce the data used in this study and present a pipeline for computing and estimating word-level AGS using a parallel dialectal corpus.

\newpage 


\section{Data and Resources}
\label{sec:dataset}

\paragraph{MADAR Corpus \& MADAR-CODA}
The MADAR corpus \cite{madar-corpus-and-lexicon} is a 26-way parallel dataset of 2,000 sentences from the Basic Travel Expression Corpus (BTEC), translated into MSA and 25 Arabic city dialects (\textbf{MADAR-26}). A subset of five dialects—CAI, BEI, DOH, TUN, and RAB—was extended with 10,000 additional sentences each (\textbf{MADAR-6}). \textbf{MADAR-CODA} \cite{madar-coda} provides CODA-normalized versions of 2,000 MADAR-6 sentences.

\paragraph{MADAR Lexicon} was also developed as part of the MADAR project, as a multilingual, multi-dialectal lexical resource, covering 1,045 concepts across 25 Arabic city dialects, along with English, French, and Modern Standard Arabic (MSA) \cite{madar-corpus-and-lexicon}. Each concept is defined using a triplet of words (En, Fr, MSA) and populated with dialectal variants annotated for both CODA orthography and CAPHI phonology. The lexicon includes 47,466 dialectal entries.

\paragraph{CAPHI Table} The CAMeL Arabic Phonetic Inventory (CAPHI) table, part of the CODA* guidelines, supports consistent, phonologically informed orthographic choices across dialects \cite{Habash:2018:unified}. It includes: (1) a CAPHI column with phonemes, (2) a CODA column with their script representations, and (3) a default mapping. For example, the CAPHI phoneme /p/ is mapped to \<ب>~\textit{b} in CODA, as in \<بري>~\textit{bry} /p~r~i/ ‘price’ (Algiers), but is assigned the default phonetic value /b/. The table draws from a large number of dialects, guiding the standardized representation of sounds absent in MSA.

\begin{table*}[t]
\centering
\tabcolsep10pt
\begin{small}
\begin{tabular}{rlcccccc}
\toprule
& \textbf{MSA} & \textbf{CAI} & \textbf{BEI} & \textbf{DOH} & \textbf{RAB} & \textbf{TUN} & \textbf{English Gloss} \\
\midrule
(1) &\<إنها>    & \<هو>       & \<هوي>     & \<موجود>     & \<كاين>     & \<موجود> & it is/ it exists    \\  
& \textit{ĂnhA}    & \textit{hw}       & \textit{hwy}     & \textit{mwjwd}     & \textit{kAyn}     & \textit{mwjwd} &    \\ \cline{2-8}

(2a) &\<في>     & \<في> & \<بأخر>      &  \<في>     & \<في>    & \<في>       &      (in) (the) end (of)   \\
& \textit{fy}     & \textit{fy} & \textit{bÂxr}       &  \textit{fy}     & \textit{fy}   & \textit{fy}       &        \\
(2b) &\<أخر>     & \<اخر>      & \<بأخر>    & \<نهاية>     & \<اللاخر>   & \<اخر>  &     \\ 
& \textit{Âxr}     & \textit{Axr}      & \textit{bÂxr}    & \textit{nhAy{\TAMAR}}     & \textit{AllAxr}   & \textit{Axr}  &       \\\cline{2-8}

(3) &\<القاعة>  & \<القاعة>   & \<الصالة>  & \<الممر>     & \<القاعة>   & \<الكولوار> & corridor  / hallway \\ 
& \textit{AlqA{\AYN}{\TAMAR}}  & \textit{AlqA{\AYN}{\TAMAR}}   & \textit{AlSAl{\TAMAR}}  & \textit{Almmr}     & \textit{AlqA{\AYN}{\TAMAR}}   & \textit{AlkwlwAr} &   \\\cline{2-8}

(4a) &\<سوف>     & \<حأجيبلك> & \<رح>      &  \<بجيب>     & ---                   & \<و>       &  I will bring you      \\
& \textit{swf}     & \textit{HÂjyblk} & \textit{rH}      &  \textit{bjyb}     & ---                   & \textit{w}       &        \\
(4b) &\<آتي>     & \<حأجيبلك> & \<جبلك>    & \<بجيب>     & \<انجيب>    & \<نجيبهولك> &   \\
& \textit{Āty}     & \textit{HÂjyblk} & \textit{jblk}    & \textit{bjyb}     & \textit{Anjyb}    & \textit{njybhwlk} &    \\
(4c) &\<لك>      & \<حأجيبلك> & \<جبلك>    & \<لك>       & \<ليك>      & \<نجيبهولك> &  \\
& \textit{lk}      & \textit{HÂjyblk} & \textit{jblk}    & \textit{lk}       & \textit{lyk}      & \textit{njybhwlk} &  \\\cline{2-8}
(5) &\<ببعض>    & \<شوية>    & \<شوي>     &  \<شوي>      & \<شويا>     & ---   &      some           \\ 
& \textit{bb{\AYN}D}    & \textit{šwy{\TAMAR}}    & \textit{šwy}     &  \textit{šwy}      & \textit{šwyA}     & ---   &                  \\
\bottomrule
\end{tabular}
\end{small}
\caption{Word-level alignments between an MSA sentence and its dialectal equivalents across five Arabic varieties.}
\label{tab:msa-dialect-alignments-complex}
\end{table*}

\section{Methodology}
\label{sec:approach}
Our methodology consists of five main components: aligning dialectal word pairs, computing a phonologically informed edit distance, aggregating these distances into a lexical similarity score, estimating this score for unseen words in context, and extending the measure from the word to the sentence level.

\subsection{Word Alignment} The task of word alignment over parallel corpora refers to identifying semantically equivalent words across sentences in different dialects or languages. 
More formally, given a set of $k$ parallel sentences 
\[
\mathbf{s} = \{s^{(d_1)}, s^{(d_2)}, \dots, s^{(d_k)}\},
\]
where each sentence is associated with a dialect label $d_i \in \mathcal{D}$ (e.g., MSA, CAI, BEI, etc.), and
\[
s^{(d)} = \langle w_1^{(d)}, w_2^{(d)}, \dots, w_{n_d}^{(d)} \rangle
\]
is a tokenized sequence in dialect $d$, the goal is to extract sets of word correspondences that capture equivalent meaning:
\[
\mathcal{A} \subseteq \left\{ \{w_p^{(d_i)}, w_q^{(d_j)}\} \;\middle|\; w_p^{(d_i)} \approx w_q^{(d_j)},\; d_i \ne d_j \right\},
\]
where $\approx$ denotes semantic equivalence between words across dialects.


We use \textbf{AWESOME Align} \cite{awesome-align}, a neural word alignment method based on multilingual contextual embeddings (see Appendix~\ref{app:AWESOME}). To handle multiple parallel sentences, we align each dialectal sentence to a central MSA anchor, assuming dialectal words aligned to the same MSA word are mutually aligned. See example in  Table~\ref{tab:msa-dialect-alignments-complex}.
More formally, let $d_1 = \text{MSA}$, and define:
\begin{multline}
\mathcal{A}_{\text{MSA}} = \left\{ \{w^{(d_1)}, w^{(d_2)}, \dots, w^{(d_k)}\} \;\middle|\right. \\
\left. w^{(d_i)} \approx w^{(d_1)},\ \forall i \in [2,k] \right\}
\end{multline}


We aggregate alignments for each word–dialect pair across the entire corpus. Given a word $w$ in dialect $d$, we define:
\[
\mathcal{A}(w, d) = \left\{ a \in \mathcal{A}_{\text{MSA}} \;\middle|\; w^{(d)} \in a \right\},
\]
where each $a \in \mathcal{A}_{\text{MSA}}$ is a set of semantically aligned words across dialects. This aggregation allows us to collect more occurrences of each word in cross-dialectal contexts, refining the signal used in our AGS modeling. 

For example, consider the MSA word \<أردت> \textit{Ârdt} `I/you wanted'. After aggregating all alignments of this word across MADAR-6, we obtain $\mathcal{A}(\text{    \<أردت>   }, \text{MSA})$ illustrated in Table~\ref{tab:aggregated-example-ardat}. Instead of just one reference, aggregation allows for multiple equivalent references in the other dialects. 

Moving forward, we will use the aggregated alignments $\mathcal{A}(w,d)$ to compute the AGS for a word $w$ in dialect $d$. Linking this back to the previous example, we will first have to compute the distance between the word \<أردت> and all its dialectal counterparts. For that, we need to define an edit distance function that is robust under the lack of standard orthography in DA.

\noindent
\begin{table*}[t]
\centering
\begin{small}
\begin{tabular}{l l}
\toprule
\textbf{Dialect} & \textbf{Aligned Forms with Frequencies} \\
\midrule

\textbf{MSA}     & \<أردت>  \textit{Ârdt}\\
\midrule

\textbf{BEI}     &
\<بدك>  \textit{bdk} (4), \<بتحب> \textit{btHb}  (1), \<بدي> \textit{bdy}  (2)  \\    
\textbf{CAI}     & \<محتاج> \textit{mHtAj}  (1), \<عايز>  \textit{{\AYN}Ayz} (5), \<عاوز> \textit{{\AYN}Awz}  (1) \\
\textbf{TUN}     & \<تستحق> \textit{tstHq}  (1), \<تحب> \textit{tHb} (3), \<نحب> \textit{nHb} (2),
\textit{None} (1) \\

\textbf{DOH}     & \<احتجت> \textit{AHtjt} (1), \<بغيت> \textit{b{\GAYN}yt} (1), \<ابغي> \textit{Ab{\GAYN}y} (2), \<بتوقف> \textit{btwqf}  (1), \<تبغي> \textit{tb{\GAYN}y} (2) \\	

\textbf{RAB}     & \<حتاجيتي> \textit{HtAjyty} (1), \<بغييتي> \textit{b{\GAYN}yyty} (1), \<نبغي> \textit{nb{\GAYN}y}  (1), \<بغيتي> \textit{b{\GAYN}yty}  (1), \<بغيت> \textit{b{\GAYN}yt}  (1), \<كنت> \textit{knt} (1), \textit{None} (1) \\
\bottomrule
\end{tabular}
\end{small}
\caption{Aggregated alignments for the MSA word \<أردت>~\textit{Ârdt} `I want' across five dialects. Numbers in parentheses indicate frequency of occurrence in the corpus.}
\label{tab:aggregated-example-ardat}
\end{table*}

\begin{table}[t]
\centering
\tabcolsep2pt
\footnotesize
\begin{tabular}{cccl}
\toprule
\textbf{CODA} & \textbf{CAPHI} & \textbf{Dialect} & \textbf{CODA-CAPHI Alignment} \\
\midrule
\<جلد>~\textit{jld} & /dj i l i d/ & KHA & [(\<ج>, dj),(-1, i),(\<ل>, l),(-1, i),(\<د>, d)] \\
\<جلد>~\textit{jld} & /g e l d/    & CAI & [(\<ج>, g),(-1, e),(\<ل>, l),(\<د>, d)] \\
\<جلد>~\textit{jld} & /j a l d/    & RAB & [(\<ج>, j),(-1, a),(\<ل>, l),(\<د>, d)] \\
\<جلد>~\textit{jld} & /j i l i d/  & BEI & [(\<ج>, j),(-1, i),(\<ل>, l),(-1, i), (\<د>, d)] \\
\<جلد>~\textit{jld} & /y i l d/    & DOH & [(\<ج>, y),(-1, i),(\<ل>, l),(\<د>, d)] \\
\bottomrule
\end{tabular}
\caption{Dialectal phonological variants of \<جلد>~\textit{jld} `leather' from the MADAR Lexicon, with letter-to-phoneme alignments.}
\label{table:madar-lex-jeem-alignments}
\end{table}


\subsection{Augmented Edit Distance} 
Consider the edit distance between \<قلب> \textit{qlb} in DOH and \<ألب>~\textit{Âlb} in BEI \textit{‘heart’}. The Levenshtein edit distance penalizes the first character substitution (\<ق> q $\leftrightarrow$ \<أ> Â), though both forms share the same etymological root \<ق>~\textit{q}. This variation stems from dialect-specific phonological realization—/g/ in DOH and /2/ in BEI—rather than true lexical divergence. To address this, we augment the Levenshtein algorithm with an etymology-aware substitution cost.\\

\vspace{-25pt}

Formally, for any character, we can define three variables representing orthography, etymology and phonology:

\verb|     |\textbf{$(x_{\ortho}, x_{\etym}, x_{\phon})$} = \textbf{(\<أ> \textit{Â} , \<ق> \textit{q}, /2/)}

\verb|     |\textbf{$(y_{\ortho}, y_{\etym}, y_{\phon})$} = \textbf{(\<ق> \textit{q} , \<ق> \textit{q}, /g/)} 

\noindent where only $x_{\ortho}, y_{\ortho}$ are observed, and the underlying $(x_{\etym}, y_{\etym})$ are equivalent. To handle such cases, we define the substitution cost as:
\begin{multline}
\label{eq:cost}
\text{cost}(x_{\text{\ortho}}, y_{\text{\ortho}} \mid d_x, d_y) = 1 - \\
P(x_{\text{\etym}} = y_{\text{\etym}} \mid x_{\text{\ortho}}, y_{\text{\ortho}}, d_x, d_y)
\end{multline}


The substitution cost of $x_{\ortho}$ with $y_{\ortho}$ in dialects $(d_x, d_y)$ is proportional to the probability they differ etymologically. Estimating this requires three components:

\begin{table}[t]
\centering
\tabcolsep3pt
\footnotesize
\begin{tabular}{cccc}
\toprule
\textbf{($x_{\etym}$, $x_{\phon}$)} & \textbf{(Phon = Etym)} & \textbf{Count} & \textbf{Etym\#} \\
\midrule
(\<ج>~\textit{j}, /dj/) & 1 & 1 & 1 \\
(\<ج>~\textit{j}, /g/)  & 0 & 1 & 1 \\
(\<ج>~\textit{j}, /j/)  & 1 & 2 & 2 \\
(\<ج>~\textit{j}, /y/)  & 0 & 1 & 1 \\
(\<ل>~\textit{l}, /l/)  & 1 & 5 & 0 \\
(\<د>~\textit{d}, /d/)  & 0 & 5 & 0 \\
\bottomrule
\end{tabular}
\caption{Etymological mappings inferred from CODA-PHON alignments in Table~\ref{table:madar-lex-jeem-alignments}.}
\label{table:jeem-caphi-summary}
\end{table}


 \paragraph{1. Phoneme from Etymology and Dialect:}

 $P(x_{\text{\phon}} \mid x_{\text{\etym}}, d_x)$. This represents the probability of a phoneme $x_{\text{\phon}}$ given its etymological character $x_{\text{\etym}}$ in dialect $d_x$. To estimate this distribution, we use the \textbf{MADAR Lexicon} \cite{madar-corpus-and-lexicon}.
We align CODA characters to CAPHI phonemes via Levenshtein alignment, guided by the CAPHI mapping table \cite{unified-DA-guidelines-caphi-coda*}, then compute the conditional probability as:
    
    {\small \[
    P(x_{\text{\phon}} \mid x_{\text{\etym}}, d_x) = \frac{\text{count}(x_{\text{\phon}}, x_{\text{\etym}}, d_x)}{\text{count}(x_{\text{\etym}}, d_x)}
    \] }

    \paragraph{2. Phoneme from Orthography and Dialect:}


$P(x_{\text{\phon}} \mid x_{\text{\ortho}}, d_x)$ is the probability of a phoneme given an orthographic character in a dialect. To estimate it, we extend the \textbf{MADAR Lexicon} with unnormalized forms from \textbf{MADAR-CODA}, aligning these to CAPHI phonemes to compute:

    \begin{small}
    \[
    P(x_{\text{\phon}} \mid x_{\text{\ortho}}, d_x) = \frac{\text{count}(x_{\text{\phon}}, x_{\text{\ortho}}, d_x)}{\text{count}(x_{\text{\ortho}}, d_x)}
    \]
    \end{small}

    \paragraph{3. Phoneme-Based Etymology Detection:}
    The orthographic character $x_{\text{\ortho}}$ can sometimes directly reflect the etymological character $x_{\text{\etym}}$, particularly when the spelling is etymologically motivated. Certain grapheme-to-phoneme mappings, such as \<ق>~\textit{q}$\rightarrow$ /2/, may serve as indicators that $x_{\ortho} = x_{\etym} $.
    We infer these etymologically consistent spellings from the \textbf{MADAR Lexicon} using the heuristic: \textit{If a character within one CODA word maps to different default and non-default CAPHI phonemes across dialects, then the orthographic form must preserve the etymology.} An illustrative example from the MADAR Lexicon is presented in Tables \ref{table:madar-lex-jeem-alignments} and \ref{table:jeem-caphi-summary}. As \<ج>~j in the CODA word \<جلد>~\textit{jld} was mapped to different default and non-default phonemes, we deduce that these mappings were etymological. We apply this rule to the MADAR Lexicon to estimate $P(x_{\etym} = x_{\ortho}| x_{\ortho}, x_{\phon}, d_x)$.

Now that all components have been defined, we can compute the posterior probability of the etymological form $x_{\text{\etym}}$ given the observed orthographic character $x_{\text{\ortho}}$ in dialect $d_x$. Using the law of total probability over possible phonemic realizations $x_{\text{\phon}} \in \text{CAPHI}$, we obtain:                                         
\vspace{-15pt}
\begin{center}
\begin{small}
\begin{multline*}  
P(x_{\text{\etym}} \mid x_{\text{\ortho}}, d_x) =\\
 \sum_{x_{\text{\phon}} \in \text{CAPHI}} 
P(x_{\text{\etym}} \mid x_{\text{\ortho}}, x_{\text{\phon}}, d_x) \cdot P(x_{\text{\phon}} \mid x_{\text{\ortho}}, d_x)
\end{multline*}
\end{small}
\end{center}

\noindent where the final term $P(x_{\text{\phon}} \mid x_{\text{\ortho}}, d_x)$ refers to the second component discussed above.
To compute the first term, we condition on etymological spelling ($x_{\ortho} = x_{\etym}$):

{\small
\[
\begin{aligned}
&P(x_{\text{\etym}} \mid x_{\text{\ortho}}, x_{\text{\phon}}, d_x)=\\
&P(x_{\text{\etym}} \mid x_{\text{\ortho}}, x_{\text{\phon}}, d_x, x_{\etym} = x_{\ortho})\cdot P(x_{\etym} = x_{\ortho} \mid x_{\text{\ortho}}, x_{\text{\phon}}, d_x) \\
& \verb|                         |+\\
&P(x_{\text{\etym}} \mid x_{\text{\ortho}}, x_{\text{\phon}}, d_x, x_{\etym} \neq x_{\ortho}) \cdot 
P(x_{\etym} \neq x_{\ortho} \mid x_{\text{\ortho}}, x_{\text{\phon}}, d_x)
\end{aligned}
\]
}

\noindent Since non-etymological spellings follow an etymology-to-phonology mapping, this reduces to:

{\small
\[
\begin{aligned}
&P(x_{\text{\etym}} \mid x_{\text{\ortho}}, x_{\text{\phon}}, d_x)=\\ 
& \mathds{1}(x_{\text{\etym}} = x_{\text{\ortho}}) \cdot 
P(x_{\etym} = x_{\ortho} \mid x_{\text{\ortho}}, x_{\text{\phon}}, d_x)\\
&+P(x_{\text{\etym}} \mid x_{\text{\phon}}, d_x) \cdot 
P(x_{\etym} \neq x_{\ortho} \mid x_{\text{\ortho}}, x_{\text{\phon}}, d_x)
\end{aligned}
\]
}



\noindent which we have prepared in the above-mentioned components 1 and 3. 
Finally, as defined before, the substitution cost between two characters $x$ and $y$ from dialects $d_x$ and $d_y$ as the probability that they \emph{do not} share the same etymological origin \ref{eq:cost}. To estimate this probability, we sum over all latent etymological character $c$ (i.e., possible CODA representations):

\vspace{-15pt}
{
\small
\begin{equation}
\begin{aligned}
&P(x_{\text{\etym}}=y_{\text{\etym}} \mid x, y, d_x, d_y)=\\
&\sum_{c \in \text{CODA}} 
P(x_{\text{\etym}} = c \mid x_{\text{\ortho}}, d_x) \cdot P(y_{\text{\etym}} = c \mid y_{\text{\ortho}}, d_y)
\end{aligned}
\label{eq:shared-etym}
\end{equation}
}

Each of the individual terms is computed as described above, by marginalizing over phonemic realizations and conditioning on whether the orthographic form is etymological. Furthermore, we can retrieve the most probable etymology for the substitution by identifying the tuples 
\( (x_{\text{\etym}}, x_{\text{\ortho}}, x_{\text{\phon}}) \) and \( (y_{\text{\etym}}, y_{\text{\ortho}}, y_{\text{\phon}}) \) 
that maximize the product in Equation~\ref{eq:shared-etym}. These tuples correspond to the most likely alignment path through the etymological space.

\subsection{Aggregating Distances into AGS}
We now aggregate the distances into a scalar AGS. Intuitively, a word is more ``general'' if it closely aligns with words from many other dialects. For each dialect, we select the \textit{minimum} distance between $w$ and any of its aligned forms from that dialect. This yields a dictionary of minimum distances $\{\delta_d\}$, where each $\delta_d$ represents the closest aligned counterpart from dialect $d$.



High distance values often reflect coincidental overlap rather than true etymological similarity—for instance, a distance of 0.8 may be no more meaningful than 1.0. To address this, we apply a smoothed threshold using a logistic function, which softly weights distances based on proximity to the cutoff, rather than applying a hard filter:

\begin{equation}
f(d) = \frac{1}{1 + e^{-s(t - d)}}
\label{eq:smooth-threshold}
\end{equation}

\noindent The function outputs values near 1 for distances well below the threshold \( t \), and near 0 for those far above. It is symmetric around \( f(t) = 0.5 \), with \( s \) controlling steepness. This enables smooth weighting of alignment quality without hard cutoffs.

The final AGS is computed by averaging the smoothed values across dialects:

\begin{equation}
\text{AGS}(w) = \frac{1}{|\mathcal{D}|} \sum_{d \in \mathcal{D}} f(\delta_d)
\label{eq:aggregator}
\end{equation}

%

We test three threshold settings with fixed steepness \( s = 20 \): 
\[
(t, s) \in \{(0.3, 20), \; (0.4, 20), \; (0.5, 20)\}
\]
Lower \( t \) values enforce stricter alignment, while higher ones permit looser matches. Fixing \( s \) ensures a sharp transition near \( t \).

\subsection{Estimating AGS}
\label{models}

In the final stage, we train a regression model to predict the AGS of a word in context. Given a word $w$ from dialect $d$, we retrieve one or more sentences from the MADAR corpus in which $w$ appears. Each sentence is pre-processed by marking the target word using dedicated special tokens. For example, if the sentence is: \<أنا متوقف عن العمل> \textit{{\AHAMZAUP}nA mtwqf {\AYN}n Al{\AYN}ml} and the target word is \<متوقف> \textit{mtwqf}, we transform the sentence to:

\<عن العمل> [TGT] \<متوقف> [TGT]  \<أنا>

\textit{{\AHAMZAUP}nA [TGT] mtwqf [TGT] {\AYN}n Al{\AYN}ml}.

These special tokens help the model focus on the word of interest and learn a mapping from its context to the AGS. We fine-tune the pretrained \textbf{CAMeL-BERT}\footnote{\url{https://huggingface.co/CAMeL-Lab/bert-base-arabic-camelbert-mix}} model \cite{inoue-etal-2021-interplay} for regression, using mean squared error (MSE) loss.







\subsection{From Word-Level to Sentence-Level AGS}
\label{Method:sentence-level}
To extend our word-level AGS to the sentence level, we propose an aggregation method that accounts for the disproportionate impact of specific words. Intuitively, the presence of just one highly specific word can drastically reduce the perceived generality of an entire sentence.

We compute the harmonic mean over the $k$ lowest-scoring words in each sentence. The harmonic mean penalizes low values more heavily than the arithmetic mean, making it well-suited for capturing the influence of specific items on overall sentence AGS. Formally, let $s = \{w_1, w_2, \dots, w_n\}$ be a sentence with $g(w_1), g(w_2), \dots, g(w_n)$ being word-level AGS's. Let $\{g_1, g_2, \dots, g_k\}$ be the $k$ lowest scores in the sentence. The sentence-level AGS $G(s)$ is computed as:
\begin{equation}
\label{eq:geom-mean}
G(s) = \frac{k}{\sum_{i=1}^{k} \frac{1}{g_i}}
\end{equation}
The choice of $k$ is treated as a tunable hyperparameter, optimized in downstream experiments.

\section{Experiments and Results}

\label{sec:results}

\subsection{AGS Statistics}
\begin{figure}[t]
    \centering
    \includegraphics[width=\linewidth]{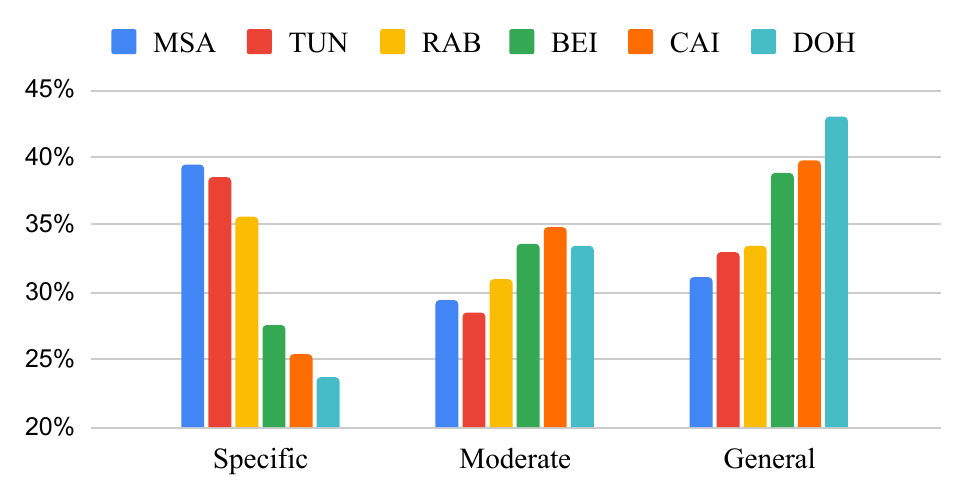}
    \caption{Distribution of Specific (0-0.1), Moderate (0.1-0.5), and General (0.5-1) annotations in AGS-annotated MADAR-6.}
    \label{fig:AGS_dist}
\end{figure}

Figure \ref{fig:AGS_dist} presents the distribution of AGS annotations, categorized as Specific (0-0.1), 
Moderate (0.1-0.5), and General (0.5-1.0) across six varieties in MADAR-6. 
MSA exhibits the highest proportion of Specific words ($\sim$39\%). In contrast, DOH (Doha) and BEI (Beirut) show a strong skew toward General words, with DOH reaching over 43\%, 
suggesting broader lexical overlap across dialects. Moderate AGS scores are more evenly distributed, with all dialects clustering around 30–34\%. The trend highlights dialectal variation in lexical generality and suggests that some dialects (e.g., DOH, BEI) may serve as better hubs for cross-dialectal generalization. This relatively low generality 
in MSA may be tied to the stylistic and structural nature of its translations. MSA forms are often longer, more formal, and semantically over-specified compared to dialectal variants. Concretely, MSA translations have the highest average sentence length in both characters 
(35.8) and words (8.0), 
compared to dialects like DOH 
(27.8 characters, 5.3 words) 
and BEI  (28.5 characters, 5.6 words). 
These inflated constructions reduce surface-level overlap with other varieties.

\subsection{Evaluation of AGS Estimation}

\paragraph{Test Set}


For evaluation, we use the MDID-DEV and MDID-TEST datasets from the NADI 2024 shared task \cite{NADI-2024}, which include 120 and 1000 Arabic tweets, respectively, annotated for dialectal validity across 11 country-level dialects. We convert the multi-label annotations into a scalar \textit{sentence-level AGS} by computing the ratio of valid dialect labels to the total number of dialects:
\begin{equation}
\label{eq:sent_level_AGS}
\text{AGS}_{\text{sent}} = \frac{n_{\text{valid}}}{n}
\end{equation}
%
%
where $n_{\text{valid}}$ is the number of dialects a sentence is annotated with, and $n$ is the total dialect count. This matches our AGS definition: broader validity implies more general vocabulary. We then apply our \textit{word-level AGS model} to each sentence by averaging predicted AGS scores across its words (Section~\ref{models}).

\paragraph{Metric}
We evaluate the predicted sentence-level scores against the ground truth AGS derived from MDID using the \textbf{Root Mean Squared Error (RMSE)} $\sqrt{\frac{1}{N} \sum_{i=1}^{N} (\hat{y}_i - y_i)^2} $, where $\hat{y}_i$ is the predicted AGS and $y_i$ is the ground truth score for sentence $i$.

\paragraph{Baselines}
We compare our method against three baselines: (X1) \textbf{MADAR Lookup}, which assigns AGS values based on a direct lookup from the annotated MADAR-26 lexicon, defaulting to 0.5 for out-of-vocabulary words; (X2) \textbf{B2BERT}, from the NADI 2024 leaderboard, which uses binary classifiers per dialect and achieves 0.5963 macro-F1 on the MDID test set;\footnote{\url{https://huggingface.co/AHAAM/B2BERT}} and (X3) the \textbf{NADI2024-baseline}, an official baseline using a top-$p$ inference strategy, which achieves 0.4697 macro-F1.\footnote{\url{https://huggingface.co/AMR-KELEG/NADI2024-baseline}} For both X2 and X3, dialect predictions are converted to sentence-level AGS by averaging over predicted dialect labels, using our aggregation method (Section~\ref{Method:sentence-level}).
\begin{table}[t]
\centering
\begin{tabular}{lc}
\toprule
\textbf{Model} & \textbf{RMSE} \\
\midrule
\multicolumn{2}{l}{\textbf{Trained Models}} \\
\midrule
CAMeL-BERT on MADAR-26 & \textbf{0.2698} \\
CAMeL-BERT on MADAR-6 & 0.2704 \\
\midrule
\multicolumn{2}{l}{\textbf{Baselines}} \\
\midrule
X1: MADAR Lookup & 0.2901 \\
X2: B2BERT & 0.3003 \\
X3: NADI2024-baseline & 0.2985 \\
\bottomrule
\end{tabular}
\caption{RMSE for AGS estimation across trained models and baselines on MDID-test.}
\label{tab:rmse}
\end{table}
\paragraph{Discussion}
Table~\ref{tab:rmse} shows that both of our AGS models—\textbf{CAMeL-BERT on MADAR-6} and \textbf{CAMeL-BERT on MADAR-26}—outperform all baselines, with the latter achieving the lowest RMSE. This suggests that finetuning specifically for AGS yields better generality estimates than MDID-based models. The small performance gap between MADAR-6 and MADAR-26 suggests that a small geographically-diverse set of dialects (as in MADAR-6) has a strong generality signal relative to a more fine-grained 26-variety setup.

{\setlength{\lineskiplimit}{0pt}\setlength{\lineskip}{0pt}
Table~\ref{tab:AGS example} illustrates how AGS captures generality in context. The sentences from the MDID dataset were annotated as valid in one and two out of 11 dialects, corresponding to sentence-level AGS scores of 0.091 and 0.273, respectively. In the first example, high-AGS words such as \<من>~\textit{mn} and \<مين>~\textit{myn} are broadly used across dialects, while lower-scoring terms like \<مخنوق>~\textit{mxnWq} and \<طايق>~\textit{TAyq} are more dialect-specific. In the second example, \<مافي>~\textit{mAfy} scores highly due to its wide regional usage, whereas expressions like \<وآخرتنا>~\textit{wAxrtnA} and \<مفر>~\textit{mfr} receive lower scores. The two least general words in each case drive the predicted AGS.
}

\begin{table}[h!t]
\centering
\tabcolsep10pt

\begin{tabular}{lll}
\multicolumn{3}{c}{\textbf{AGS:} 0.091 ,  \textbf{Predicted AGS:} 0.173}\\
\toprule
\textbf{Word} & \textbf{Gloss} & \textbf{AGS} \\ \midrule
\<مين>~\textit{myn} & who & 0.986 \\
\<هيكون>~\textit{hykwn} & will be & 0.725 \\
\<جنبك>~\textit{jnbk} & beside you & 0.817 \\
\<ف>~\textit{f} & in & 0.941 \\
\<حزنك>~\textit{Hznk} & your sadness & 0.549 \\
\<وزعلك>~\textit{wz{\AYN}lk} & and your upset & 0.448 \\
\<مين>~\textit{myn} & who & 0.986 \\
\<هيكون>~\textit{hykwn} & will be & 0.725 \\
\<جنبك>~\textit{jnbk} & beside you & 0.817 \\
\<وانت>~\textit{wAnt} & while you are & 0.637 \\
\<مخنوق>~\textit{mxnwq} & suffocating & \cellcolor{yellow}0.229 \\
\<ومش>~\textit{wm{\SHIN}} & and not & 0.371 \\
\<طايق>~\textit{TAyq} & tolerating & \cellcolor{yellow}0.139 \\
\<الدنيا>~\textit{AldnyA} & the world & 0.498 \\
\bottomrule
\multicolumn{3}{c}{ }\\
\end{tabular}
\begin{tabular}{lll}
\multicolumn{3}{c}{\textbf{AGS:} 0.273 ,  \textbf{Predicted AGS:} 0.278}\\
\toprule
\textbf{Word} & \textbf{Gloss} & \textbf{Score} \\ \midrule
\<مافي>~\textit{mAfy} & there is no & 0.821 \\
\<مفر>~\textit{mfr} & escape & \cellcolor{yellow}0.312 \\
\<من>~\textit{mn} & from & 0.987 \\
\<هالشغله>~\textit{hAl{\SHIN}glh} & this thing & 0.525 \\
\<واخرتنا>~\textit{wAxrtnA} & and eventually & \cellcolor{yellow}0.251 \\
\<بدنا>~\textit{bdnA} & we want & 0.465 \\
\<نكمل>~\textit{nkml} & to complement & 0.812 \\
\<نص>~\textit{nS} & half & 0.943 \\
\<ديننا>~\textit{dynnA} & our religion & 0.431 \\
\bottomrule
\end{tabular}
\caption{Two MDID sentences with word-level AGS predicted by CAMeL-BERT (MADAR-26). Sentence-level AGS is computed as the geometric mean~\ref{eq:geom-mean} of the two least-general (k=2) highlighted words. }
\label{tab:AGS example}
\end{table}

\vspace{-15pt}
\section{Conclusions and Future Work}
We introduced the Arabic Generality Score (AGS), a new measure of how widely a word is used across dialects, complementing existing metrics like ALDi. We built an annotation pipeline and trained a contextual model that outperforms strong baselines on a multi-dialect benchmark.

Future work includes extending AGS to phrases and constructions, modeling variation across domains and time, and integrating AGS into downstream tasks like translation, retrieval, and educational NLP.




\section*{Limitations}
While AGS offers a promising step toward multidimensional modeling of Arabic dialectness, it has several limitations. First, our method depends on the availability of parallel dialectal corpora, which are still scarce and often skewed toward certain regions such as the Levant or Egypt. This limits the diversity of dialectal features in the data.

Second, the annotation pipeline assumes that surface similarity aligns with functional or semantic equivalence across dialects, which is not always the case, particularly with idioms or culturally specific expressions.

Third, the model may overfit to frequent patterns and underperform on low-resource varieties. We also do not currently include geographic or sociolinguistic metadata, which may help improve AGS estimation.

\section*{Ethics Statement}
Our work aims to support more inclusive, dialect-aware Arabic NLP. However, dialect modeling can raise concerns related to identity, marginalization, and language hierarchies. AGS is a descriptive tool, not a value judgment of language use.

We used only publicly available corpora and followed standard ethical practices. Still, dialectal data may reflect existing biases, including underrepresentation of certain regions or groups. We encourage future efforts to broaden dialect coverage.

Finally, AGS models could be misused for profiling or surveillance. We strongly oppose any application of this work in ways that compromise privacy or reinforce discrimination.

We used AI writing assistance within the scope of ``Assistance
purely with the language of the paper'' described in the ACL Policy on Publication Ethics.


\bibliographystyle{acl_natbib}
\bibliography{custom,camel-bib-v3,anthology}

\begin{thebibliography}{27}
\providecommand{\natexlab}[1]{#1}

\bibitem[{Abdul-Mageed et~al.(2023)Abdul-Mageed, Elmadany, Zhang, Nagoudi, Bouamor, and Habash}]{NADI-2023}
Muhammad Abdul-Mageed, AbdelRahim Elmadany, Chiyu Zhang, El~Moatez~Billah Nagoudi, Houda Bouamor, and Nizar Habash. 2023.
\newblock \href {https://doi.org/10.18653/v1/2023.arabicnlp-1.62} {{NADI} 2023: The fourth nuanced {A}rabic dialect identification shared task}.
\newblock In \emph{Proceedings of ArabicNLP 2023}, pages 600--613, Singapore (Hybrid). Association for Computational Linguistics.

\bibitem[{Abdul-Mageed et~al.(2024)Abdul-Mageed, Keleg, Elmadany, Zhang, Hamed, Magdy, Bouamor, and Habash}]{NADI-2024}
Muhammad Abdul-Mageed, Amr Keleg, AbdelRahim Elmadany, Chiyu Zhang, Injy Hamed, Walid Magdy, Houda Bouamor, and Nizar Habash. 2024.
\newblock \href {https://doi.org/10.18653/v1/2024.arabicnlp-1.79} {{NADI} 2024: The fifth nuanced {A}rabic dialect identification shared task}.
\newblock In \emph{Proceedings of The Second Arabic Natural Language Processing Conference}, pages 709--728, Bangkok, Thailand. Association for Computational Linguistics.

\bibitem[{Abdul-Mageed et~al.(2020)Abdul-Mageed, Zhang, Bouamor, and Habash}]{NADI-2020}
Muhammad Abdul-Mageed, Chiyu Zhang, Houda Bouamor, and Nizar Habash. 2020.
\newblock \href {https://aclanthology.org/2020.wanlp-1.9/} {{NADI} 2020: The first nuanced {A}rabic dialect identification shared task}.
\newblock In \emph{Proceedings of the Fifth Arabic Natural Language Processing Workshop}, pages 97--110, Barcelona, Spain (Online). Association for Computational Linguistics.

\bibitem[{Abdul-Mageed et~al.(2021)Abdul-Mageed, Zhang, Elmadany, Bouamor, and Habash}]{NADI-2021}
Muhammad Abdul-Mageed, Chiyu Zhang, AbdelRahim Elmadany, Houda Bouamor, and Nizar Habash. 2021.
\newblock \href {https://aclanthology.org/2021.wanlp-1.28/} {{NADI} 2021: The second nuanced {A}rabic dialect identification shared task}.
\newblock In \emph{Proceedings of the Sixth Arabic Natural Language Processing Workshop}, pages 244--259, Kyiv, Ukraine (Virtual). Association for Computational Linguistics.

\bibitem[{Abdul-Mageed et~al.(2022)Abdul-Mageed, Zhang, Elmadany, Bouamor, and Habash}]{NADI-2022}
Muhammad Abdul-Mageed, Chiyu Zhang, AbdelRahim Elmadany, Houda Bouamor, and Nizar Habash. 2022.
\newblock \href {https://doi.org/10.18653/v1/2022.wanlp-1.9} {{NADI} 2022: The third nuanced {A}rabic dialect identification shared task}.
\newblock In \emph{Proceedings of the Seventh Arabic Natural Language Processing Workshop (WANLP)}, pages 85--97, Abu Dhabi, United Arab Emirates (Hybrid). Association for Computational Linguistics.

\bibitem[{Alhafni et~al.(2024)Alhafni, Al-Towaity, Fawzy, Nassar, Eryani, Bouamor, and Habash}]{codafication}
Bashar Alhafni, Sarah Al-Towaity, Ziyad Fawzy, Fatema Nassar, Fadhl Eryani, Houda Bouamor, and Nizar Habash. 2024.
\newblock \href {https://doi.org/10.18653/v1/2024.arabicnlp-1.4} {Exploiting dialect identification in automatic dialectal text normalization}.
\newblock In \emph{Proceedings of The Second Arabic Natural Language Processing Conference}, pages 42--54, Bangkok, Thailand. Association for Computational Linguistics.

\bibitem[{Badawi and Hinds(1986)}]{badawi1986dictionary}
El-Said~M. Badawi and Martin Hinds. 1986.
\newblock \emph{A Dictionary of Egyptian Arabic}.
\newblock Librairie du Liban, Beirut.

\bibitem[{Badawi(1973)}]{badawi1973mustawayat}
Said~M. Badawi. 1973.
\newblock \emph{Mustaway{\={a}}t al-{\c{c}}arabiyya al-mu{\={a}}{\c{s}}ira f{\={\i}} Mi{\d{s}}r}.
\newblock D{\={a}}r al-Ma{\c{c}}{\={a}}rif bi-Mi{\d{s}}r, Cairo.

\bibitem[{Bouamor et~al.(2018)Bouamor, Habash, Salameh, Zaghouani, Rambow, Abdulrahim, Obeid, Khalifa, Eryani, Erdmann, and Oflazer}]{madar-corpus-and-lexicon}
Houda Bouamor, Nizar Habash, Mohammad Salameh, Wajdi Zaghouani, Owen Rambow, Dana Abdulrahim, Ossama Obeid, Salam Khalifa, Fadhl Eryani, Alexander Erdmann, and Kemal Oflazer. 2018.
\newblock \href {https://aclanthology.org/L18-1535/} {The {MADAR} {A}rabic dialect corpus and lexicon}.
\newblock In \emph{Proceedings of the Eleventh International Conference on Language Resources and Evaluation ({LREC} 2018)}, Miyazaki, Japan. European Language Resources Association (ELRA).

\bibitem[{Dou and Neubig(2021)}]{awesome-align}
Zi-Yi Dou and Graham Neubig. 2021.
\newblock \href {https://doi.org/10.18653/v1/2021.eacl-main.181} {Word alignment by fine-tuning embeddings on parallel corpora}.
\newblock In \emph{Proceedings of the 16th Conference of the European Chapter of the Association for Computational Linguistics: Main Volume}, pages 2112--2128, Online. Association for Computational Linguistics.

\bibitem[{Eryani et~al.(2020)Eryani, Habash, Bouamor, and Khalifa}]{madar-coda}
Fadhl Eryani, Nizar Habash, Houda Bouamor, and Salam Khalifa. 2020.
\newblock \href {https://aclanthology.org/2020.lrec-1.508/} {A spelling correction corpus for multiple {A}rabic dialects}.
\newblock In \emph{Proceedings of the Twelfth Language Resources and Evaluation Conference}, pages 4130--4138, Marseille, France. European Language Resources Association.

\bibitem[{Eskander et~al.(2013)Eskander, Habash, Rambow, and Tomeh}]{Eskander:2013:processing}
Ramy Eskander, Nizar Habash, Owen Rambow, and Nadi Tomeh. 2013.
\newblock Processing spontaneous orthography.
\newblock In \emph{Proceedings of the Conference of the North American Chapter of the Association for Computational Linguistics (NAACL)}, pages 585--595, Atlanta, Georgia.

\bibitem[{Ferguson(1959)}]{ferguson1959diglossia}
Charles~A. Ferguson. 1959.
\newblock Diglossia.
\newblock \emph{Word}, 15(2):325--340.

\bibitem[{Habash et~al.(2018{\natexlab{a}})Habash, Eryani, Khalifa, Rambow, Abdulrahim, Erdmann, Faraj, Zaghouani, Bouamor, Zalmout, Hassan, Al-Shargi, Alkhereyf, Abdulkareem, Eskander, Salameh, and Saddiki}]{unified-DA-guidelines-caphi-coda*}
Nizar Habash, Fadhl Eryani, Salam Khalifa, Owen Rambow, Dana Abdulrahim, Alexander Erdmann, Reem Faraj, Wajdi Zaghouani, Houda Bouamor, Nasser Zalmout, Sara Hassan, Faisal Al-Shargi, Sakhar Alkhereyf, Basma Abdulkareem, Ramy Eskander, Mohammad Salameh, and Hind Saddiki. 2018{\natexlab{a}}.
\newblock \href {https://aclanthology.org/L18-1574/} {Unified guidelines and resources for {A}rabic dialect orthography}.
\newblock In \emph{Proceedings of the Eleventh International Conference on Language Resources and Evaluation ({LREC} 2018)}, Miyazaki, Japan. European Language Resources Association (ELRA).

\bibitem[{Habash et~al.(2018{\natexlab{b}})Habash, Khalifa, Eryani, Rambow, Abdulrahim, Erdmann, Faraj, Zaghouani, Bouamor, Zalmout, Hassan, shargi, Alkhereyf, Abdulkareem, Eskander, Salameh, and Saddiki}]{Habash:2018:unified}
Nizar Habash, Salam Khalifa, Fadhl Eryani, Owen Rambow, Dana Abdulrahim, Alexander Erdmann, Reem Faraj, Wajdi Zaghouani, Houda Bouamor, Nasser Zalmout, Sara Hassan, Faisal~Al shargi, Sakhar Alkhereyf, Basma Abdulkareem, Ramy Eskander, Mohammad Salameh, and Hind Saddiki. 2018{\natexlab{b}}.
\newblock {Unified Guidelines and Resources for Arabic Dialect Orthography}.
\newblock In \emph{Proceedings of the International Conference on Language Resources and Evaluation (LREC 2018)}, Miyazaki, Japan.

\bibitem[{Habash et~al.(2008)Habash, Rambow, Diab, and Faraj}]{dialectness_annotation_guidelines}
Nizar Habash, Owen Rambow, Mona Diab, and Reem Faraj. 2008.
\newblock Guidelines for annotation of arabic dialectness.
\newblock \emph{Proceedings of the LREC Workshop on HLT \& NLP within the Arabic world}.

\bibitem[{Habash et~al.(2007)Habash, Soudi, and Buckwalter}]{Habash:2007:arabic-transliteration}
Nizar Habash, Abdelhadi Soudi, and Tim Buckwalter. 2007.
\newblock {On {{A}rabic} Transliteration}.
\newblock In A.~van~den Bosch and A.~Soudi, editors, \emph{{A}rabic Computational Morphology: Knowledge-based and Empirical Methods}, pages 15--22. Springer, Netherlands.

\bibitem[{Hamed et~al.(2020)Hamed, Vu, and Abdennadher}]{hamed-etal-2020-arzen}
Injy Hamed, Ngoc~Thang Vu, and Slim Abdennadher. 2020.
\newblock \href {https://aclanthology.org/2020.lrec-1.523} {{A}rz{E}n: A speech corpus for code-switched {E}gyptian {A}rabic-{E}nglish}.
\newblock In \emph{Proceedings of the Twelfth Language Resources and Evaluation Conference}, pages 4237--4246, Marseille, France. European Language Resources Association.

\bibitem[{Inoue et~al.(2021)Inoue, Alhafni, Baimukan, Bouamor, and Habash}]{inoue-etal-2021-interplay}
Go~Inoue, Bashar Alhafni, Nurpeiis Baimukan, Houda Bouamor, and Nizar Habash. 2021.
\newblock \href {https://aclanthology.org/2021.wanlp-1.10} {The interplay of variant, size, and task type in {A}rabic pre-trained language models}.
\newblock In \emph{Proceedings of the Sixth Arabic Natural Language Processing Workshop}, pages 92--104, Kyiv, Ukraine (Virtual). Association for Computational Linguistics.

\bibitem[{Iriarte~Diez et~al.(2023)Iriarte~Diez, Laaber, Kampen, and Fernández}]{diez-white-arabic}
Ana Iriarte~Diez, Claudia Laaber, Nina Kampen, and Monserrat Fernández. 2023.
\newblock \href {https://doi.org/10.31810/rsel.53.2.9} {What is white arabic? new labels in a changing arab world}.
\newblock \emph{Revista Española de Lingüística}, 53:229--266.

\bibitem[{Keleg et~al.(2023)Keleg, Goldwater, and Magdy}]{ALDI}
Amr Keleg, Sharon Goldwater, and Walid Magdy. 2023.
\newblock \href {https://doi.org/10.18653/v1/2023.emnlp-main.655} {{ALD}i: Quantifying the {A}rabic level of dialectness of text}.
\newblock In \emph{Proceedings of the 2023 Conference on Empirical Methods in Natural Language Processing}, pages 10597--10611, Singapore. Association for Computational Linguistics.

\bibitem[{Keleg and Magdy(2023)}]{limitations-of-single-label-DI}
Amr Keleg and Walid Magdy. 2023.
\newblock \href {https://doi.org/10.18653/v1/2023.arabicnlp-1.31} {{A}rabic dialect identification under scrutiny: Limitations of single-label classification}.
\newblock In \emph{Proceedings of ArabicNLP 2023}, pages 385--398, Singapore (Hybrid). Association for Computational Linguistics.

\bibitem[{Olsen et~al.(2023)Olsen, Touileb, and Velldal}]{madar_error_analysis}
Helene Olsen, Samia Touileb, and Erik Velldal. 2023.
\newblock \href {https://doi.org/10.18653/v1/2023.arabicnlp-1.30} {{A}rabic dialect identification: An in-depth error analysis on the {MADAR} parallel corpus}.
\newblock In \emph{Proceedings of ArabicNLP 2023}, pages 370--384, Singapore (Hybrid). Association for Computational Linguistics.

\bibitem[{Sajjad et~al.(2020)Sajjad, Abdelali, Durrani, and Dalvi}]{Arabench-DA-Eng-MT-benchmark}
Hassan Sajjad, Ahmed Abdelali, Nadir Durrani, and Fahim Dalvi. 2020.
\newblock \href {https://doi.org/10.18653/v1/2020.coling-main.447} {{A}ra{B}ench: Benchmarking dialectal {A}rabic-{E}nglish machine translation}.
\newblock In \emph{Proceedings of the 28th International Conference on Computational Linguistics}, pages 5094--5107, Barcelona, Spain (Online). International Committee on Computational Linguistics.

\bibitem[{Salameh et~al.(2018)Salameh, Bouamor, and Habash}]{fine-grained-DI-madar}
Mohammad Salameh, Houda Bouamor, and Nizar Habash. 2018.
\newblock \href {https://aclanthology.org/C18-1113/} {Fine-grained {A}rabic dialect identification}.
\newblock In \emph{Proceedings of the 27th International Conference on Computational Linguistics}, pages 1332--1344, Santa Fe, New Mexico, USA. Association for Computational Linguistics.

\bibitem[{Zaidan and Callison-Burch(2011)}]{AOC}
Omar~F. Zaidan and Chris Callison-Burch. 2011.
\newblock \href {https://aclanthology.org/P11-2007/} {The {A}rabic online commentary dataset: an annotated dataset of informal {A}rabic with high dialectal content}.
\newblock In \emph{Proceedings of the 49th Annual Meeting of the Association for Computational Linguistics: Human Language Technologies}, pages 37--41, Portland, Oregon, USA. Association for Computational Linguistics.

\bibitem[{Zaidan and Callison-Burch(2014)}]{DI-long-paper-zaidan-2014}
Omar~F. Zaidan and Chris Callison-Burch. 2014.
\newblock \href {https://doi.org/10.1162/COLI_a_00169} {{A}rabic dialect identification}.
\newblock \emph{Computational Linguistics}, 40(1):171--202.

\end{thebibliography}

\onecolumn
\appendix
\section{AWESOME Align}
\label{app:AWESOME}
Given a pair of sequences 
\[
x = \langle x_1, \dots, x_n \rangle \quad \text{and} \quad y = \langle y_1, \dots, y_m \rangle,
\]
AWESOME \cite{awesome-align} extracts contextual embeddings
\[
h_x \in \mathbb{R}^{n \times d}, \quad h_y \in \mathbb{R}^{m \times d},
\]
from a shared encoder and computes a similarity matrix:
\[
S = h_x h_y^\top.
\]
A row-wise softmax yields the alignment probability matrix:
\[
S_{xy}^{(i,j)} = \frac{\exp(h_{x_i} \cdot h_{y_j})}{\sum_{j'=1}^m \exp(h_{x_i} \cdot h_{y_{j'}})}.
\]
Alignments are extracted by applying a symmetric agreement criterion: a word pair $(x_i, y_j)$ is aligned if the forward and backward probabilities exceed a confidence threshold $\tau$:
\[
(x_i, y_j) \in \mathcal{A} \quad \text{iff} \quad S_{xy}^{(i,j)} > \tau \quad \text{and} \quad S_{yx}^{(j,i)} > \tau.
\]

We finetuned the encoder CAMeLBERT \cite{inoue-etal-2021-interplay} using all of AWESOME's proposed objectives: (1) Masked Language Modeling (MLM) ,  applied independently on source and target sentences, (2) Translation Language Modeling (TLM) ,  MLM on concatenated source–target pairs to encourage cross-sentence alignment, (3) Self-Training Objective (SO) ,  promotes closeness of initially aligned token embeddings, (4) Parallel Sentence Identification (PSI) ,  contrastive loss distinguishing parallel from non-parallel sentences, and (5) Consistency Optimization (CO) ,  maximizes agreement between forward and backward alignment matrices. We finetuned CAMeLBERT on approximately 100{,}000 MSA–DA sentence pairs from the MADAR-26 corpus, covering 25 dialects. Although MADAR contains no gold-standard alignments, previous work has shown that these objectives significantly improve alignment accuracy.

\section{Implementation}

\paragraph{Hyperparameters}
For the aggregation function~\ref{eq:smooth-threshold}, we evaluated three threshold values\\ $t \in \{0.3, 0.4, 0.5\}$ and selected $t = 0.5$ based on the lowest RMSE for AGS estimation on the MDID-DEV set. For contextual AGS prediction, we fine-tune CAMeL-BERT using a batch size of 32, a learning rate of $4 \times 10^{-5}$ with a linear decay schedule and zero warmup steps, and the AdamW optimizer. Training was capped at 10{,}000 steps with early stopping based on validation performance. All experiments were run with a fixed random seed of 42.

\paragraph{Hardware}
All experiments were conducted on Google Colab using an NVIDIA A100 GPU. Training checkpoints were saved every 1,000 steps and monitored using Weights~\&~Biases.

\paragraph{Libraries and Tools}
We used Python with \texttt{pandas}, \texttt{numpy}, and \texttt{camel-tools} for data preprocessing and orthographic normalization. Sentence alignment relied on \texttt{AWESOME-Align}, and contextual modeling was implemented with \texttt{transformers (v4.x)} and \texttt{PyTorch}.










\end{document}